\PassOptionsToPackage{table}{xcolor}

\documentclass[sigconf]{acmart}

\usepackage{graphicx}
\usepackage[table]{xcolor}
\usepackage{algorithm}
\usepackage{algorithmic}
\usepackage{afterpage}
\usepackage{placeins}
\usepackage{array}
\usepackage{cuted}
\usepackage{listings}
\lstdefinestyle{codeappendix}{
  language=Python,
  basicstyle=\ttfamily\footnotesize,
  keywordstyle=\bfseries,
  commentstyle=\itshape\color{gray},
  breaklines=true,
  breakatwhitespace=false,
  showstringspaces=false,
  columns=fullflexible,
  keepspaces=true,
  tabsize=4,
}
\copyrightyear{2026}
\acmYear{2026}
\setcopyright{cc}
\setcctype{by}
\acmConference[MM '26]{Proceedings of the 34th ACM International Conference on Multimedia}{November 10--14, 2026}{Rio de Janeiro, Brazil}
\acmBooktitle{Proceedings of the 34th ACM International Conference on Multimedia (MM '26), November 10--14, 2026, Rio de Janeiro, Brazil}
\acmDOI{10.1145/3767308.3836126}
\acmISBN{979-8-4007-2213-4/2026/11}
\begin{document}

\title[Coarse-to-Fine Two-Step VLA]{CF-VLA: Efficient Coarse-to-Fine Action Generation for Vision-Language-Action Policies}


\author{Fan Du}
\orcid{0009-0006-3433-0966}
\authornote{Equal contribution.}
\authornote{Work done during internships at Enacta AI.}
\affiliation{
  \institution{Southern University of Science and Technology}
  \city{Shenzhen}
  \country{China}
}

\author{Feng Yan}
\authornotemark[1]
\authornotemark[2]
\authornote{National Key Laboratory of Human-Machine Hybrid Augmented Intelligence, National Engineering Research Center for Visual Information and Applications, and Institute of Artificial Intelligence and Robotics, Xi'an Jiaotong University}
\orcid{0009-0004-6267-9423}
\affiliation{
  \institution{Xi’an Jiaotong University}
  \city{Xi'an}
  \country{China}
}
\email{bphengyan@163.com}

\author{Jianxiong Wu}
\orcid{0000-0001-9608-2255}
\authornotemark[2]
\affiliation{
  \institution{Enacta AI}
  \city{Shenzhen}
  \country{China}
}

\author{Xinrun Xu}
\orcid{0000-0001-8162-8702}
\authornotemark[2]
\affiliation{
  \institution{University of Chinese Academy of Sciences}
  \city{Beijing} 
  \country{China}
}

\author{Weiye Zhang}
\orcid{0009-0001-1177-1850}
\authornotemark[2]
\affiliation{
  \institution{Enacta AI}
  \city{Shenzhen}
  \country{China}
}

\author{Weinong Wang}
\orcid{0000-0001-9731-6631}
\authornotemark[3]
\affiliation{
  \institution{Xi’an Jiaotong University}
  \city{Xi'an}
  \country{China}
}

\author{Yu Guo}
\orcid{0000-0002-5489-8288}
\authornotemark[3]
\affiliation{
  \institution{Xi’an Jiaotong University}
  \city{Xi'an}
  \country{China}
}

\author{Bin Qian}
\orcid{0009-0001-3968-5882}
\authornotemark[2]
\affiliation{
  \institution{Tsinghua University}
  \city{Shenzhen}
  \country{China}
}

\author{Zhihai He}
\orcid{0000-0002-2647-8286}
\affiliation{
  \institution{Southern University of Science and Technology}
  \city{Shenzhen}
  \country{China}
}

\author{Fei Wang}
\authornotemark[3]
\orcid{0000-0003-3462-8472}
\affiliation{
  \institution{Xi’an Jiaotong University}
  \city{Xi’an}
  \country{China}
}
\email{wfx@mail.xjtu.edu.cn}

\author{Heng Yang}
\orcid{0000-0002-7798-0049}
\affiliation{
  \institution{Enacta AI}
  \city{Shenzhen}
  \country{China}
}
\email{heng.yang@enacta.ai}
\renewcommand{\shortauthors}{Fan Du et al.}


\begin{abstract}

Flow-based vision-language-action (VLA) policies offer strong expressivity for action generation, but suffer from a fundamental inefficiency: multi-step inference is required to recover action structure from uninformative Gaussian noise, leading to a poor efficiency–quality trade-off under real-time constraints. We address this issue by rethinking the role of the starting point in generative action modeling. Instead of shortening the sampling trajectory, we propose \textbf{CF-VLA}, a coarse-to-fine two-stage formulation that restructures action generation into a coarse initialization step that constructs an action-aware starting point, followed by a single-step local refinement that corrects residual errors. Concretely, the coarse stage learns a conditional posterior over endpoint velocity to transform Gaussian noise into a structured initialization, while the fine stage performs a fixed-time refinement from this initialization. To stabilize training, we introduce a stepwise strategy that first learns a controlled coarse predictor and then performs joint optimization. Experiments on CALVIN and LIBERO show that our method establishes a strong efficiency–performance frontier under low-NFE (Number of Function Evaluations) regimes: it consistently outperforms existing NFE=2 methods, matches or surpasses the NFE=10 $\pi_{0.5}$ baseline on several metrics, reduces action sampling latency by 73.2\%, and achieves the best hard-task real-robot success rate of 74.7\%, outperforming MIP by 26.0 points and reaching comparable or slightly better performance than $\pi_{0.5}$ with a fivefold smaller inference budget. 
These results suggest that structured, coarse-to-fine generation enables both strong performance and efficient inference. Our code is available at \url{https://github.com/EmbodiedAI-RoboTron/CF-VLA}.
\end{abstract}


\maketitle

\begin{figure*}[t]
  \centering
  \includegraphics[width=0.992\textwidth]{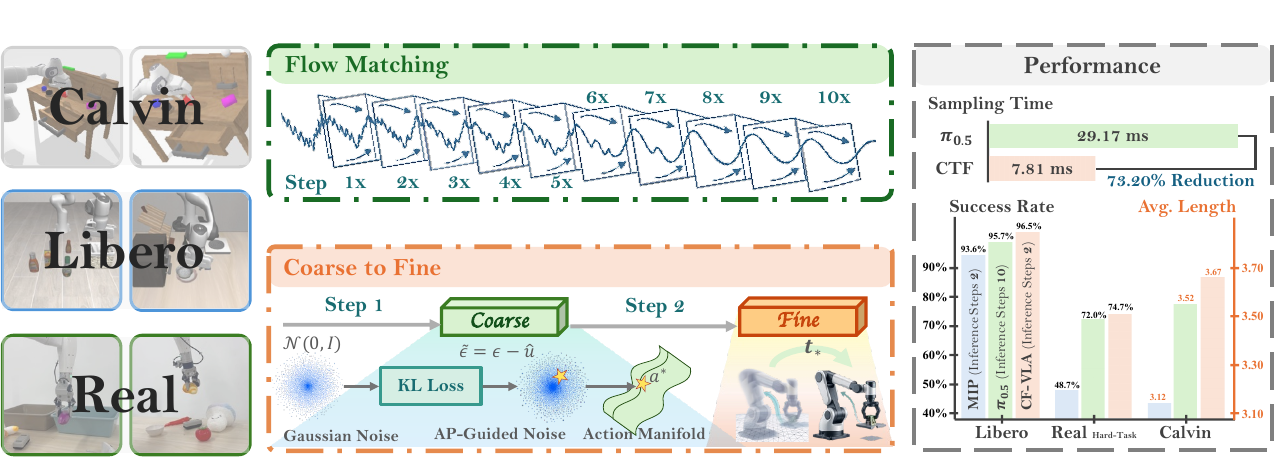}
  \caption{Teaser of CF-VLA. Standard flow matching requires multiple iterative steps to recover action structure from uninformative Gaussian noise. CF-VLA instead adopts a coarse-to-fine two-step process: a coarse stage constructs an action-prior-guided (AP-guided) noise initialization, followed by a single-step refinement. This design achieves a stronger efficiency–performance frontier across CALVIN, LIBERO, and real-robot settings, reducing action sampling latency by 73.2\%.}
  \Description{A teaser figure of CF-VLA across CALVIN, LIBERO, and real-robot settings. The left column shows representative tasks from the three evaluation domains. The upper middle panel illustrates standard flow matching iteratively refining from Gaussian noise over many steps. The lower middle panel shows the proposed coarse-to-fine design, which first uses a KL-supervised coarse stage to transform Gaussian noise into action-prior-guided (AP-guided) noise initialization and then uses a fine stage to recover the final action in one step. The right panel summarizes the resulting latency reductions and performance gains.}
  \label{fig:teaser}
\end{figure*}

\section{Introduction}
Flow-based VLA policies~\cite{arxiv2504.16054,arxiv2510.07077,arxiv2508.13073,arxiv2410.24164} have recently emerged as a promising direction for robotic manipulation, 
with methods such as $\pi_{0.5}$ offering strong expressivity for modeling multimodal continuous actions.
However, their practical deployment remains limited by a fundamental inefficiency: 
action generation starts from uninformative Gaussian noise and requires multiple iterative refinement steps to recover meaningful structure, 
leading to a poor efficiency--quality trade-off under real-time constraints.


A useful perspective on this limitation comes from the recent analysis of Liu et al.~\cite{arxiv2512.02826}, who study oracle velocity in flow-based diffusion models and show that the target dynamics exhibit a two-stage structure: early-time evolution mainly performs navigation from the Gaussian prior toward relevant modes, whereas later-time dynamics become increasingly dominated by the nearest data sample and focus on refining fine-grained details. This observation helps explain why aggressively reducing the number of sampling steps can make high-quality generation difficult: a small inference budget forces global transport and local correction to compete within the same truncated trajectory. Consistent with this view, recent work in generative robotic control~\cite{arxiv2512.01809} suggests that splitting inference into two MSE-supervised stages—a coarse, large-step update followed by a smaller corrective refinement—can already achieve performance on par with flow-based policies $\pi_{0}$~\cite{arxiv2410.24164}.

In this work, we revisit flow-based action generation from a coarse-to-fine perspective and argue that efficient generation 
requires restructuring the starting point rather than shortening the trajectory. 
We decompose the generation process into two explicit stages: 
(i) a coarse initialization stage that resolves the global mismatch between Gaussian noise and the action manifold by constructing an AP-guided initialization, 
and (ii) a fine refinement stage that performs a single-step local correction to recover the final action. 
Concretely, the coarse stage learns a conditional posterior over endpoint velocity to transform Gaussian noise into an AP-guided initialization, 
while the fine stage performs a fixed-time refinement from this initialization. 
To stabilize training, we further introduce a stepwise optimization strategy that first learns a controlled coarse predictor 
and then switches to full joint training, ensuring that the refinement stage operates in a well-conditioned local regime. 
As illustrated in Figure~\ref{fig:teaser}, this design replaces iterative refinement from unstructured noise with a two-step pipeline 
of AP-guided initialization followed by local correction.

Our method achieves a favorable balance between action quality and efficiency under a strict two-step inference budget. 
On LIBERO~\cite{arxiv2306.03310} and CALVIN~\cite{arxiv2112.03227}, it matches or surpasses our reproduced $\pi_{0.5}$ baseline while significantly reducing computation, 
including a 73.2\% reduction in action sampling latency. 
We further validate the method on real-robot manipulation, where CF-VLA achieves the best hard-task average success rate of 74.7\%, outperforming MIP by 26.0 points under the same NFE=2 budget and reaching comparable or slightly better performance than $\pi_{0.5}$ with a fivefold smaller inference budget.
These results suggest that the efficiency bottleneck of flow-based policies lies in the structure of the starting point, 
rather than the length of the sampling trajectory.

Overall, our contributions are threefold: 
\begin{itemize}
     \item[(1)] A coarse-to-fine two-stage VLA framework that explicitly separates global alignment and local refinement; the framework is plug-and-play and can be seamlessly applied to arbitrary flow-based VLA policies.
     \item[(2)] A variance-aware endpoint distribution formulation that transforms Gaussian noise into AP-guided initializations closer to the valid action manifold;
     \item[(3)] A stepwise training strategy that stabilizes cross-stage coupling and enables efficient low-step generation.
    
\end{itemize}

\section{Coarse-to-Fine Two-Step Action Generation}
\subsection{Overview}

\begin{figure*}[t]
  \centering
  \includegraphics[width=\linewidth]{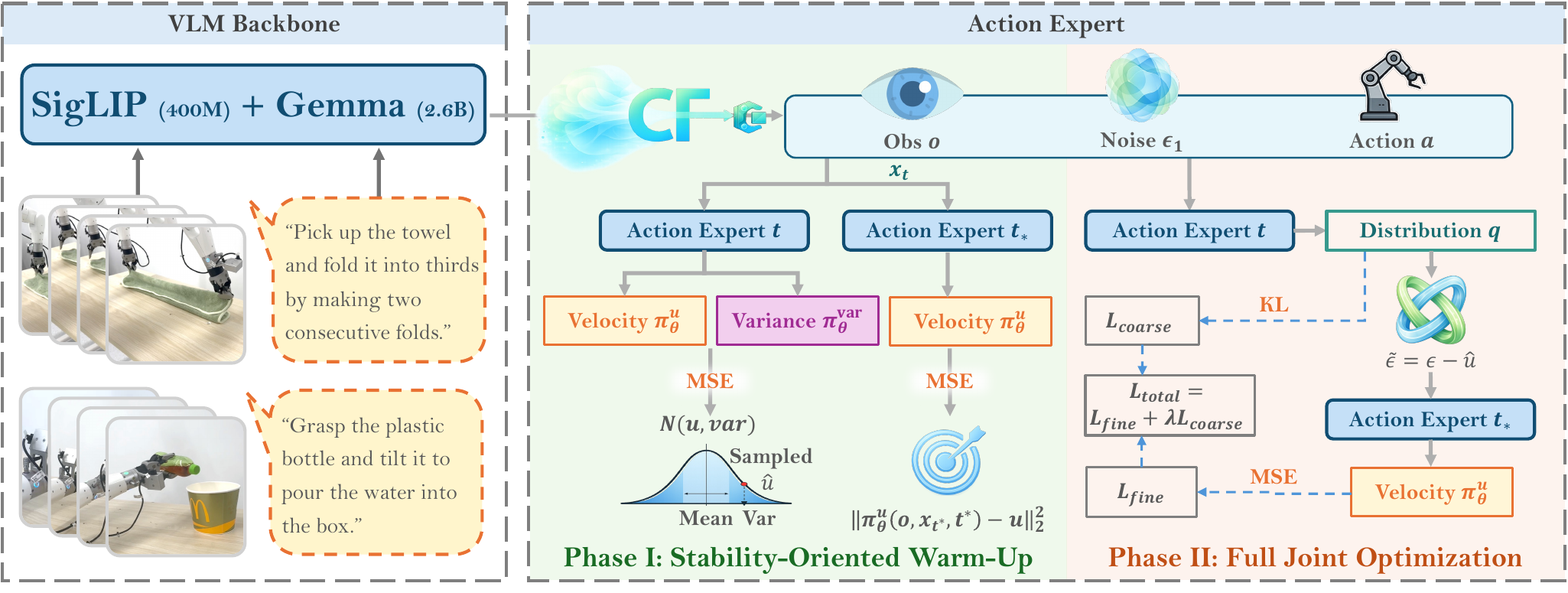}
  \caption{Overview of CF-VLA. CF-VLA adopts a two-phase training strategy. Phase I is a stability-oriented warm-up stage that shapes endpoint velocity and variance prediction and trains the refinement branch on a controlled proxy input, avoiding unreliable coarse outputs at the beginning of optimization. Phase II then performs full joint optimization of the final coarse-to-fine mechanism: KL-supervised endpoint posterior matching transforms pure Gaussian noise into an AP-guided initialization, and a single-step refinement branch performs local correction to recover the target action. This two-phase design enables stable training and efficient two-step generation.}
  \Description{A method overview of CF-VLA. A VLM backbone provides conditioning features to a shared action expert. The diagram contains two training phases. Phase I serves as a stabilization phase, where the endpoint branch learns a stable velocity and variance representation together with a controlled refinement proxy to provide reliable coarse outputs. Phase II then performs coarse-to-fine generation, where the endpoint branch predicts a posterior matched by KL divergence to produce AP-guided noise initialization, and the fine branch predicts a mean-squared-error-supervised update that maps the local neighborhood induced by this AP-guided initialization onto the ground-truth action manifold.}
  \label{fig:overview}
\end{figure*}

\begin{figure}[t]
  \centering
  \includegraphics[width=\columnwidth]{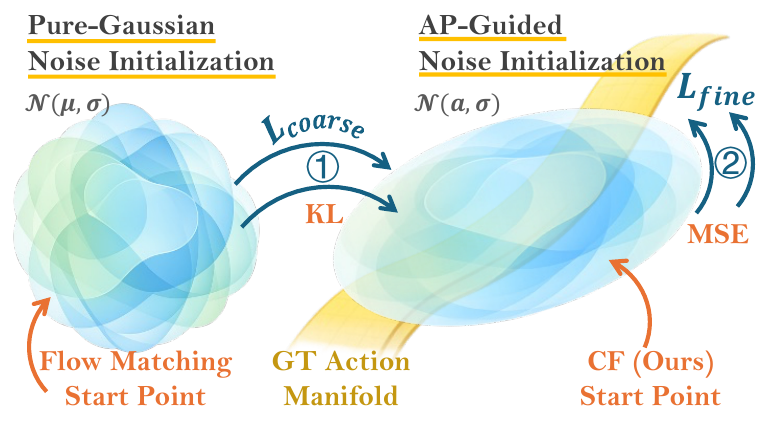}
  \caption{Geometric view of CF-VLA. Standard flow matching starts from pure Gaussian noise, forcing early steps to spend computation on global transport toward the task-conditioned action manifold. CF-VLA instead first builds an AP-guided initialization distribution with a KL-supervised coarse stage, then applies a single refinement stage to recover the ground-truth action.}
  \Description{A conceptual diagram comparing pure Gaussian noise initialization and AP-guided initialization relative to the ground-truth action manifold. The coarse loss with KL divergence reshapes the initial distribution toward a local neighborhood around the target manifold induced by the AP-guided initialization, and the fine loss with mean squared error maps this neighborhood onto the target manifold.}
  \label{fig:manifold_view}
\end{figure}

We consider behavioral cloning for VLA policy learning from a dataset $\mathcal{D}=\{(o_i, a_i)\}_{i=1}^N$, where each observation $o$ consists of visual inputs, a language instruction, and the robot state, and $a \in \mathbb{R}^{H \times d}$ denotes an action chunk of horizon $H$ with dimension $d$.
As a starting point, we adopt standard flow matching (FM) to model the conditional action distribution~\cite{arxiv2210.02747,arxiv2410.24164}.

\textbf{Limitation of standard flow matching.}
Standard flow matching (FM)~\cite{arxiv2210.02747} defines a time-dependent velocity field $v_\theta(o, x_t, t)$ along the linear interpolation
\begin{equation}
    x_t = t\epsilon + (1-t)a, \qquad u_t = \epsilon - a,
\end{equation}
where $\epsilon \sim \mathcal{N}(0, I)$ and $t \in [0,1]$.

At inference time, actions are obtained by integrating this velocity field from noise to action through multiple Euler steps.

While expressive, this formulation introduces a structural inefficiency. The induced dynamics are inherently heterogeneous: early-time evolution primarily performs large-scale transport from Gaussian noise toward the action distribution, whereas late-time evolution focuses on local refinement near valid actions~\cite{arxiv2512.02826}. However, both regimes are parameterized by a single velocity field and executed through a homogeneous iterative solver~\cite{arxiv2210.02747,arxiv1806.07366}. As a result, the model repeatedly spends computation rediscovering coarse action structure from uninformative noise, even though this global alignment is largely shared across samples. Under strict inference budgets, this leads to a suboptimal efficiency--quality trade-off.

\textbf{From global transport to AP-guided initialization.}
We attribute this inefficiency to a mismatch between the structure of the problem and the structure of the computation induced by expert demonstrations~\cite{arxiv2602.11236,arxiv2407.19681}. 
As illustrated in Figure~\ref{fig:manifold_view}, standard FM ignores this structure at initialization and reconstructs it through iterative transport at every rollout. To address this, we shift the focus from shortening the sampling trajectory to restructuring the starting point. Instead of repeatedly discovering the action manifold during inference, we propose to directly initialize generation in a neighborhood already aligned with it.

Accordingly, our method decomposes action generation into two functional stages, as shown in Figure~\ref{fig:overview}:

\begin{itemize}
    \item \textbf{Coarse initialization:} construct an AP-guided initialization distribution aligned with the action manifold;
    \item \textbf{Fine refinement:} perform a single local correction to recover the final action.
\end{itemize}


Concretely, given an observation $o$, we first sample Gaussian noise $\epsilon_1 \sim \mathcal{N}(0, I)$. The coarse stage transforms $\epsilon_1$ into an AP-guided initialization $\tilde{\epsilon}$ that is biased toward the action manifold. The fine stage then applies a single refinement step from $\tilde{\epsilon}$ to predict the final action.

This design fundamentally changes low-NFE generation. Standard FM uses multiple steps to both discover a plausible action direction and refine it. In contrast, our method separates these roles: global alignment is handled once through initialization, while the limited computation budget is reserved for precise local correction.

Yet training this decomposition from scratch can be unstable: early coarse predictions are disordered and Gaussian-like, making them poor refinement inputs. Phase I (Figure~\ref{fig:overview}) introduces a stability-oriented warm-up stage, shaping endpoint velocity and variance, and supervising refinement on a proxy input rather than immature coarse outputs.

\subsection{Endpoint Prior Modeling (Coarse Stage)}

The goal of the coarse stage is not to predict the final action, but to shape the geometry of the AP-guided initialization distribution.

For each training pair $(o,a)$, we sample endpoint noise $\epsilon_1 \sim \mathcal{N}(0,I)$ and define the endpoint target velocity
\begin{equation}
u_{t_1} = \epsilon_1 - a.
\end{equation}

We model a conditional posterior over this velocity:
\begin{equation}
q_\theta(u \mid o,\epsilon_1)
=
\mathcal{N}\!\bigl(\pi_\theta^u(o,\epsilon_1,1),\; \pi_\theta^{\mathrm{var}}(o,\epsilon_1,1)\, I\bigr).
\end{equation}

This posterior is matched to the target distribution:
\begin{equation}
p(u \mid o,a,\epsilon_1)
=
\mathcal{N}(u_{t_1}, \sigma_{\mathrm{noise}}^{2} I)
\end{equation}
where $\sigma_{\mathrm{noise}}^{2}$ is a tunable hyperparameter controlling the isotropic variance. We then optimize the KL objective
\begin{equation}
\mathcal{L}_{\text{coarse}}(\theta)
=
\mathbb{E}_{(o,a),\,\epsilon_1}
\bigl[
\mathrm{KL}\bigl(
q_\theta(u \mid o,\epsilon_1)
\;\|\;
p(u \mid o,a,\epsilon_1)
\bigr)
\bigr].
\end{equation}

This formulation explicitly treats the coarse stage as a distribution shaping problem rather than point regression. The objective is to learn a stochastic, task-conditioned AP-guided initialization distribution whose mass is concentrated near plausible actions, rather than collapsing to a deterministic estimate.

The AP-guided initialization is then defined as
\begin{equation}
\tilde{\epsilon} = \epsilon_1 - \hat{u}, \qquad \hat{u} \sim q_\theta(u \mid o,\epsilon_1).
\end{equation}
Compared with pure Gaussian noise, the mean of $\tilde{\epsilon}$ is already shifted toward the action manifold, providing a more informative starting point for subsequent refinement. This reduces the burden on the refinement stage, which no longer needs to recover global action structure from an uninformative Gaussian start.

\subsection{Local Manifold Recovery (Fine Stage)}

Once the AP-guided initialization distribution is reshaped, the problem faced by the model changes fundamentally. The fine stage no longer needs to discover global action structure; instead, it operates in a local neighborhood around the target manifold.

We therefore reformulate refinement as a local recovery problem. Given $\tilde{\epsilon}$, we apply a single update at a fixed time $t_f$:
\begin{equation}
\mathcal{L}_{\text{fine}}(\theta)
=
\mathbb{E}_{(o,a),\,\epsilon_1,\,\hat{u}}
\Bigl[
\bigl\|
\pi_\theta^u(o,\tilde{\epsilon},t_f)-(\tilde{\epsilon}-a)
\bigr\|_2^2
\Bigr].
\end{equation}

This objective is appropriate because, under a well-shaped initialization, the residual error lies in a locally smooth regime around the action manifold. In this regime, a single first-order update is sufficient to recover high-quality actions.

This decomposition converts generation from a global transport problem into a local correction problem conditioned on a learned initialization.

\subsection{Joint Objective}

The overall objective combines the two roles:
\begin{equation}
\mathcal{L}(\theta)
=
\mathcal{L}_{\text{fine}}(\theta)
+
\lambda\,\mathcal{L}_{\text{coarse}}(\theta).
\end{equation}

These two terms operate at different levels. The coarse loss shapes the geometry of the AP-guided initialization distribution, while the fine loss learns how to recover the final action from this AP-guided initialization. Removing either component breaks the decomposition: without coarse shaping, refinement must rediscover global structure; without refinement, the model cannot land on the action manifold.

\begin{algorithm}[!h]
\caption{Stepwise training of CF-VLA}
\label{alg:training}
\small
\begin{algorithmic}[1]
\REQUIRE Dataset $\mathcal{D}=\{(o,a)\}$, warm-up weight $\alpha$, joint weight $\lambda$, fixed refinement state $t_f$
\STATE Initialize model parameters $\theta$

\STATE \textbf{Phase I: Stabilization warm-up}
\REPEAT
    \STATE Sample a mini-batch $\{(o_i,a_i)\}_{i=1}^{B}$ and noise $\epsilon_1 \sim \mathcal{N}(0,I)$
    \STATE Construct the controlled proxy input $x_{t_f}=t_f\epsilon_1+(1-t_f)a$
    \STATE Compute endpoint warm-up loss for velocity and variance prediction at $(o,\epsilon_1,1)$
    \STATE Compute refinement proxy loss at $(o,x_{t_f},t_f)$
    \STATE Update $\theta$ by minimizing
    \[
    \mathcal{L}_{\mathrm{warm}}
    =
    \mathcal{L}_{\mathrm{endpoint}}
    +
    \alpha\,\mathcal{L}_{\mathrm{proxy}}
    \]
\UNTIL{warm-up converges}

\STATE \textbf{Phase II: Joint coarse-to-fine optimization}
\REPEAT
    \STATE Sample a mini-batch $\{(o_i,a_i)\}_{i=1}^{B}$ and noise $\epsilon_1 \sim \mathcal{N}(0,I)$
    \STATE Compute the endpoint target velocity $u_{t_1}=\epsilon_1-a$
    \STATE Predict the endpoint posterior $q_\theta(u \mid o,\epsilon_1)$
    \STATE Compute the coarse loss $\mathcal{L}_{\text{coarse}}$ by KL matching to the target distribution
    \STATE Sample $\hat{u}\sim q_\theta(u \mid o,\epsilon_1)$ and form $\tilde{\epsilon}=\epsilon_1-\hat{u}$
    \STATE Use $\tilde{\epsilon}$ as the fine-stage input and compute $\mathcal{L}_{\text{fine}}$ at $(o,\tilde{\epsilon},t_f)$
    \STATE Update $\theta$ by minimizing
    \[
    \mathcal{L}
    =
    \mathcal{L}_{\text{fine}}
    +
    \lambda\,\mathcal{L}_{\text{coarse}}
    \]
\UNTIL{convergence}
\STATE \textbf{return} trained parameters $\theta$
\end{algorithmic}
\end{algorithm}

\begin{table*}[!h]
  \caption{LIBERO simulation comparison with prior methods grouped by inference type and NFE. In the Aux. Mech. column, ``-'' indicates no explicit auxiliary mechanism. Other labels denote auxiliary mechanism categories including Plan, WM, and Backbone Params. $^{*}$ indicates our reproductions trained on replay-filtered LIBERO data.}
  \label{tab:libero_sota_nfe}
  \centering
    \begin{tabular}{@{}p{0.19\textwidth}p{0.14\textwidth}p{0.12\textwidth}p{0.04\textwidth}ccccc@{}}
    \toprule
    Method & Venue, Year & Aux. Mech. & NFE & Spatial & Object & Goal & Long & Avg. \\
    \midrule
    \multicolumn{9}{@{}l}{\textit{Autoregressive methods}} \\
    UniVLA~\cite{univla2025} & RSS, 2025 & 7B & AR & 96.5 & 96.8 & 95.6 & 92.0 & 95.2 \\
    UnifiedVLA~\cite{unifiedvla2026} & ICLR, 2026 & WM+8.5B & AR & 95.4 & \textbf{98.8} & 93.6 & 94.0 & 95.5 \\
    OpenVLA-OFT (7B)~\cite{kim2025finetuningvisionlanguageactionmodelsoptimizing} & RSS, 2025 & 7B & AR & 97.6 & 98.4 & 97.9 & 94.5 & 97.1 \\
    VLA-Cache (7B)~\cite{vlacache2025} & NeurIPS, 2025 & 7B & AR & \textbf{98.3} & 97.5 & \textbf{98.3} & \textbf{95.4} & \textbf{97.4} \\
    \midrule
    \multicolumn{9}{@{}l}{\textit{NFE = 1 methods}} \\
    PI-VLA~\cite{pivla2026} & Symmetry, 2026 & Plan+WM+7B & 1 & 79.5 & 73.4 & 73.3 & 66.6 & 73.2 \\
    UniAct~\cite{uniact2025} & CVPR, 2025 & - & 1 & 77.0 & 87.0 & 77.0 & 70.0 & 77.8 \\
    MolmoAct-7B-D~\cite{molmoact2025} & arXiv, 2025 & Plan+7B & 1 & 87.0 & 95.4 & 87.6 & 77.2 & 86.6 \\
    NinA (MLP)~\cite{nina2025} & \shortstack[l]{NeurIPS WS, 2025} & - & 1 & 87.8 & \textbf{98.2} & 90.2 & \textbf{92.8} & 90.9 \\
    EveryDayVLA~\cite{everydayvla2026} & \shortstack[l]{ICRA subm., 2026} & Plan+7B & 1 & \textbf{96.8} & 95.6 & \textbf{91.0} & 82.0 & \textbf{91.4} \\
    \midrule
    \multicolumn{9}{@{}l}{\textit{NFE = 8/10 methods}} \\
    DreamVLA~\cite{dreamvla2025} & NeurIPS, 2025 & WM & 10 & 97.5 & 94.0 & 89.5 & 89.5 & 92.6 \\
    $\pi_{0.5}^{*}$~\cite{arxiv2504.16054} & CoRL, 2025 & 3B & 10 & 98.4 & 97.4 & 96.2 & 90.6 & 95.7 \\
    InstructVLA~\cite{instructvla2026} & ICLR, 2026 & - & 10 & 97.3 & 99.6 & 96.5 & 89.8 & 95.8 \\
    MemoryVLA~\cite{memoryvla2026} & ICLR, 2026 & 7B & 10 & 98.4 & 98.4 & 96.4 & 93.4 & 96.5 \\
    Flower VLA~\cite{flower2025} & CoRL, 2025 & - & 8 & 97.5 & 99.1 & 96.1 & 94.9 & 96.9 \\
    $\pi_{0.5}$~\cite{arxiv2504.16054} & CoRL, 2025 & 3B & 10 & \textbf{98.8} & 98.2 & 98.0 & 92.4 & 96.9 \\
    X-VLA~\cite{xvla2026} & ICLR, 2026 & - & 10 & 98.2 & 98.6 & 97.8 & \textbf{97.6} & 98.1 \\
    Cosmos Policy~\cite{cosmos2026} & ICLR, 2026 & Plan+WM & 10 & 98.1 & \textbf{100.0} & \textbf{98.2} & \textbf{97.6} & \textbf{98.5} \\
    \midrule
    \multicolumn{9}{@{}l}{\textit{NFE = 20/100 methods}} \\
    Dita~\cite{dita2025} & ICCV, 2025 & - & 20 & 84.2 & 96.3 & 85.4 & 63.8 & 82.4 \\
    CronusVLA~\cite{cronusvla2026} & AAAI, 2026 & 7B & 100 & \textbf{97.3} & \textbf{99.6} & \textbf{96.9} & \textbf{94.0} & \textbf{97.0} \\
    \midrule
    \multicolumn{9}{@{}l}{\textit{NFE = 2 methods}} \\
    \shortstack[l]{MIP ($\pi_{0}$ arch.)~\cite{arxiv2512.01809}} & ICLR, 2026 & 3B & 2 & 97.6 & 95.8 & 95.2 & 82.2 & 92.7 \\
    \shortstack[l]{MIP ($\pi_{0.5}$ arch.)$^{*}$~\cite{arxiv2512.01809}} & ICLR, 2026 & 3B & 2 & 96.6 & 97.0 & 95.8 & 85.0 & 93.6 \\
    \shortstack[l]{$\pi_{0.5}^{*}$ (NFE=2)~\cite{arxiv2504.16054}} & CoRL, 2025 & 3B & 2 & 97.2 & 93.6 & \textbf{98.0} & 90.4 & 94.8 \\
    CF-VLA (ours) & - & 3B & 2 & \textbf{98.0} & \textbf{99.2} & \textbf{96.6} & \textbf{92.0} & \textbf{96.5} \\
    \bottomrule
  \end{tabular}
\end{table*}

\begin{table*}[!htbp]
  \caption{CALVIN long-horizon comparison with prior methods grouped by NFE. In the Aux. Mech. column, ``-'' indicates no explicit auxiliary mechanism. Other labels summarize auxiliary mechanisms such as Plan, WM, 3D spatial input, and Distill. Avg. Len. denotes the average sequence length. $^{*}$ indicates results reproduced by us.}
  \label{tab:calvin_sota_nfe}
  \centering
\begin{tabular}{@{}p{0.20\textwidth}p{0.10\textwidth}p{0.10\textwidth}ccccccc@{}}
    \toprule
    Method & Venue, Year & Aux. Mech. & NFE & 1 & 2 & 3 & 4 & 5 & Avg. Len. \\
    \midrule
    \multicolumn{10}{@{}l}{\textit{NFE = 1 methods}} \\
    DeeR~\cite{deer2024} & NeurIPS, 2024 & - & 1 & 85.3 & 69.6 & 54.9 & 42.0 & 31.2 & 2.83 \\
    LCD~\cite{lcd2024} & ICLR, 2024 & Plan & 1 & 88.7 & 69.9 & 54.5 & 42.7 & 32.2 & 2.88 \\
    DySL-VLA~\cite{dyslvla2026} & DAC, 2026 & - & 1 & 89.4 & 71.9 & 53.9 & 42.0 & 32.0 & 2.89 \\
    TaKSIE~\cite{taksie2025} & WACV, 2025 & Plan & 1 & 90.4 & 73.9 & 61.7 & 51.2 & 40.8 & 3.18 \\
    HULC++~\cite{arxiv2210.01911} & ICRA, 2023 & Plan & 1 & 93.0 & 79.0 & 64.0 & 52.0 & 40.0 & 3.30 \\
    RoboTron-Mani~\cite{robotronmani2025} & arXiv, 2025 & 3D input & 1 & 94.7 & 80.3 & 65.1 & 51.4 & 39.0 & 3.31 \\
    DaDu-Corki-SW~\cite{daducorki2025} & ISCA, 2025 & - & 1 & 92.3 & 80.0 & 67.4 & 56.6 & 45.8 & 3.42 \\
    RoboUniView (default)~\cite{robouniview2024} & arXiv, 2024 & - & 1 & \textbf{95.4} & \textbf{82.7} & 68.5 & 56.4 & 46.1 & 3.49 \\
    DTP~\cite{dtp2025} & RA-L, 2025 & Plan & 1 & 92.4 & 81.9 & \textbf{70.2} & \textbf{60.3} & \textbf{50.9} & \textbf{3.55} \\
    \midrule
    \multicolumn{10}{@{}l}{\textit{NFE = 4 methods}} \\
    LaDi-WM~\cite{ladiwm2025} & arXiv, 2025 & WM & $\approx$4 & 92.7 & 83.1 & 72.1 & 61.2 & 54.1 & 3.63 \\
    LightDP~\cite{lightdp2025} & ICCV, 2025 & Distill & 4 & \textbf{93.7} & \textbf{84.5} & \textbf{74.1} & \textbf{64.4} & \textbf{55.6} & \textbf{3.72} \\
    \midrule
    \multicolumn{10}{@{}l}{\textit{NFE $\geq$ 10 methods}} \\
    RoboTron-Mani (DiT)~\cite{robotronmani2025} & arXiv, 2025 & 3D input & $>$10 & \textbf{96.9} & \textbf{83.0} & 68.1 & 56.5 & 46.8 & 3.51 \\
    $\pi_{0.5}^{*}$~\cite{arxiv2504.16054} & CoRL, 2025 & 3B & 10 & 90.4 & 78.1 & 67.9 & \textbf{61.2} & \textbf{54.7} & 3.52 \\
    MDT (default)~\cite{mdt2024} & RSS, 2024 & - & 10 & 93.3 & 82.4 & \textbf{71.5} & 60.9 & 51.1 & \textbf{3.59} \\
    \midrule
    \multicolumn{10}{@{}l}{\textit{NFE = 2 methods}} \\
    MIP ($\pi_{0.5}$ arch.)$^{*}$~\cite{arxiv2512.01809} & ICLR, 2026 & 3B & 2 & 86.7 & 72.1 & 59.6 & 51.2 & 42.4 & 3.12 \\
    $\pi_{0.5}^{*}$ (NFE=2)~\cite{arxiv2504.16054} & CoRL, 2025 & 3B & 2 & 88.7 & 76.0 & 66.2 & 59.7 & 52.4 & 3.43 \\
    CF-VLA (ours) & - & 3B & 2 & \textbf{91.1} & \textbf{80.2} & \textbf{71.8} & \textbf{66.2} & \textbf{57.3} & \textbf{3.67} \\
    \bottomrule
\end{tabular}
\end{table*}

\subsection{Stepwise Training}

The coarse-to-fine formulation introduces an asymmetric dependency: the effectiveness of the fine stage relies on the quality of the coarse initialization. Direct joint optimization is therefore unstable.

\textbf{Phase I: Stabilization.}
We first train the endpoint representation under controlled supervision, while exposing the refinement branch to a stable proxy distribution constructed by interpolating between noise and ground-truth actions. This phase does not yet realize the full coarse-to-fine mechanism, but ensures that both branches operate on well-conditioned inputs.

\textbf{Phase II: Joint optimization.}
Once the coarse outputs become sufficiently structured, we switch to the full objective. At this stage, the refinement branch receives inputs generated by the learned coarse posterior, and joint optimization learns how distribution shaping and local recovery reinforce each other.

This stepwise strategy follows directly from the structure of the problem, rather than being a heuristic training trick.

\subsection{Algorithm Summary}
For clarity, Algorithm~\ref{alg:training} summarizes the stepwise training procedure while omitting formula expansions already defined above. In Phase I, we use the warm-up objective $\mathcal{L}_{\mathrm{warm}}=\mathcal{L}_{\mathrm{endpoint}}+\alpha\,\mathcal{L}_{\mathrm{proxy}}$, where $\mathcal{L}_{\mathrm{endpoint}}$ supervises endpoint velocity and variance prediction and $\mathcal{L}_{\mathrm{proxy}}$ supervises the refinement branch on the controlled proxy input. Inference directly follows the forward path of Phase II: Gaussian noise is first transformed into AP-guided noise initialization through posterior sampling, and one refinement update is then applied to produce the final action.

\section{Experiments}

We evaluate CF-VLA from three complementary perspectives: (i) task success in simulation, (ii) inference efficiency, and (iii) generalization to real-robot settings. 
We first describe the training and evaluation setup, then present comparisons with prior methods on LIBERO and CALVIN under varying inference budgets, together with detailed ablation studies, followed by analysis of action-generation latency and real-robot results.

\textbf{Training and evaluation setup.}
We train all models on 8 NVIDIA A100 GPUs using PyTorch FSDP, with 2 samples per GPU (global batch size 16) and a learning rate of $5 \times 10^{-5}$. 
For CALVIN, we follow the standard long-horizon D$\rightarrow$D protocol. 
For LIBERO, we replay-filter trajectories within each suite and retain successful trajectories for unified training, evaluating a checkpoint trained for 60{,}000 steps. 
Each simulation task is evaluated with 10 trials. 
For real-robot experiments, the number of trials is task-dependent and reported per task. 
We report per-suite average success rates.

\textbf{Baselines.}
We compare CF-VLA with strong generative policy baselines. 
Here, NFE denotes the number of iterative updates required at inference time. 
Rather than treating NFE as a mere implementation detail, we explicitly adopt it as a primary evaluation axis that reflects the amount of iterative corrective computation.

$\pi_{0.5}$~\cite{arxiv2504.16054} is a strong flow-matching VLA policy with iterative sampling and serves as the main performance reference. 
MIP~\cite{arxiv2512.01809} also follows a two-stage sampling pipeline and is therefore the most directly comparable low-NFE baseline. 
For fair comparison, all baselines marked with $^{*}$ are reproduced using the same training data and aligned settings. 
We further include representative low-NFE, high-NFE, and autoregressive methods to position CF-VLA across different efficiency regimes.

Notably, several high-performing baselines rely on additional auxiliary mechanisms, such as explicit planning modules (Plan), world models (WM), explicit 3D geometric inputs, or consistency distillation, as well as larger backbones, which increase system complexity and computational cost. In contrast, CF-VLA improves only the core action-generation process without introducing such extra modules, making the comparison more directly isolate the effect of the action-generation design. 

\subsection{Simulation Results}

\textbf{Benchmarks.}
We evaluate on two standard language-conditioned manipulation benchmarks: LIBERO and CALVIN. 
LIBERO measures multi-task generalization across four Franka suites (Spatial, Object, Goal, Long), while CALVIN evaluates long-horizon instruction following under the D$\rightarrow$D setting, reporting success rates for completing 1 to 5 consecutive instructions and the average sequence length~\cite{arxiv2112.03227,arxiv2204.06252,arxiv2210.01911}.

Tables~\ref{tab:libero_sota_nfe} and \ref{tab:calvin_sota_nfe} compare CF-VLA with prior methods under heterogeneous NFE budgets. 
Lower-NFE methods rely heavily on the structural quality of their initialization, whereas higher-NFE methods compensate for unstructured starting points via extended iterative refinement. 
This distinction directly reflects the role of initialization in efficient action generation.

\textbf{Results on LIBERO.}
CF-VLA demonstrates strong performance under strict low-NFE constraints. 
Under identical replay-filtered training settings, CF-VLA achieves an average success rate of 96.5 at NFE=2, outperforming the reproduced NFE=10 $\pi_{0.5}$ baseline (95.7). 
This indicates that improving the structure of the starting point enables strong performance with substantially fewer refinement steps.
Notably, this gain is achieved despite a smaller iterative budget, indicating that the benefit does not come merely from reducing computation but from allocating computation more effectively.

The gains are concentrated on Object (+1.8) and Long (+1.4), which are particularly sensitive to structured and temporally consistent action generation. This observation is consistent with our design: by injecting action priors at initialization, CF-VLA reduces the burden of global alignment during inference and allows the limited refinement budget to focus on local correction.

Within the NFE=2 regime, CF-VLA also outperforms both the reproduced MIP baseline and the NFE=2 $\pi_{0.5}$ variant, establishing a strong efficiency--performance trade-off. 
While a performance gap remains compared to high-NFE methods such as Cosmos Policy (NFE=10), these methods typically rely on additional components or higher computational budgets. 
In contrast, CF-VLA achieves competitive results by restructuring the generation process itself, rather than increasing iterative computation.

\textbf{Results on CALVIN.}
The advantage of CF-VLA becomes more pronounced as the task horizon increases. 
Compared with both the reproduced MIP baseline and the NFE=2 $\pi_{0.5}$ model, CF-VLA consistently achieves higher success rates across all 1-to-5 instruction settings. 
In particular, it improves the 5-instruction success rate by 4.9 points (57.3 vs 52.4) and increases the average sequence length from 3.43 to 3.67.

More importantly, CF-VLA exhibits strong performance even when compared across different NFE regimes. 
Compared to the best NFE=1 model, CF-VLA improves the 5-instruction success rate by 6.4 points. 
It also achieves the highest success rates on the 4- and 5-instruction settings (66.2 and 57.3) among methods with significantly higher computational budgets, such as MDT (NFE=10) and LightDP (NFE=4).

These results support our core hypothesis: separating global alignment from local refinement allows the model to operate in a well-conditioned regime, reducing error accumulation over long horizons. 
By initializing generation near the action manifold, CF-VLA enables the limited refinement budget to be spent on maintaining rollout consistency, rather than repeatedly correcting large deviations.
\begin{table}
  \caption{LIBERO simulation success rates (\%). We report the main comparison and ablation variants of CF-VLA. $^{*}$ indicates our reproductions trained on replay-filtered data.}
  \label{tab:libero_results}
  \centering
  \begin{tabular}{@{}lccccc@{}}
    \toprule
    Method & Spatial & Object & Goal & Long & Avg. \\
    \midrule
    MIP ($\pi_{0}$ arch.) & 97.6 & 95.8 & 95.2 & 82.2 & 92.7 \\
    MIP ($\pi_{0.5}$ arch.)$^{*}$ & 96.6 & 97.0 & 95.8 & 85.0 & 93.6 \\
    $\pi_{0.5}^{*}$ & \textbf{98.4} & 97.4 & \textbf{96.2} & 90.6 & 95.7 \\
    CF-VLA (full) & 98.0 & \textbf{99.2} & \textbf{96.6} & \textbf{92.0} & \textbf{96.5} \\
    \midrule
    CF-VLA w/o Phase II & 96.4 & 98.4 & 95.8 & 87.8 & 94.6 \\
    CF-VLA w/o refine. & 95.6 & \textbf{99.0} & 94.2 & 90.6 & 94.9 \\
    CF-VLA w/o var. mod. & 97.4 & 98.4 & \textbf{96.4} & 88.4 & 95.2 \\
    CF-VLA w/o Phase I & \textbf{98.4} & \textbf{99.0} & 95.0 & \textbf{90.8} & \textbf{95.8} \\
    \bottomrule
  \end{tabular}
\end{table}



\subsection{Ablation Studies}

Tables~\ref{tab:libero_results} and \ref{tab:calvin_results} present ablations of CF-VLA on both LIBERO and CALVIN. Specifically, ``w/o Phase I'' removes the stabilization warm-up and directly trains the full objective, ``w/o Phase II'' reports the Phase-I-only checkpoint, ``w/o var. mod.'' removes variance prediction in the coarse stage, and ``w/o refine.'' removes the fine recovery step. Together, these variants isolate the roles of Phase-I stabilization, Phase-II joint coarse-to-fine optimization, variance modeling, and fine refinement.

\textbf{Results on LIBERO.}
The full model achieves the best average success rate of 96.5, outperforming MIP ($\pi_{0}$ arch.) by 3.8 points, MIP ($\pi_{0.5}$ arch.) by 2.9 points, and $\pi_{0.5}^{*}$ by 0.8 points. Improvements are particularly evident on the Long suite (92.0 vs 90.6), which requires stable multi-step action consistency.

Removing Phase I leads to a modest drop to 95.8, suggesting that the stabilization stage primarily improves optimization robustness rather than being the sole source of the final gains. 
In contrast, removing refinement, removing variance modeling, or removing Phase II degrades performance more significantly, reducing the average success rate to 94.9, 95.2, and 94.6, respectively. 
Notably, removing variance modeling leads to a pronounced drop on Long (92.0 $\rightarrow$ 88.4), indicating that capturing uncertainty in the coarse stage is critical for maintaining stable long-horizon behavior.

These results align with our design: Phase I stabilizes the shared representation, Phase II couples coarse posterior prediction to fine recovery, the coarse stage shapes an AP-guided initialization distribution, and the fine stage performs local correction. 
Without Phase II, the refinement branch never learns on coarse posterior samples; without refinement, the model cannot reliably recover the final action from the AP-guided initialization; without variance modeling, the coarse stage collapses toward deterministic predictions, reducing the diversity and robustness of the initialization.

\begin{table}
  \caption{CALVIN long-horizon success rates (\%). We report both the main comparison and ablation variants of CF-VLA. $^{*}$ indicates results reproduced by us.}
  \label{tab:calvin_results}
  \centering
  \begin{tabular}{@{}lcccccc@{}}
    \toprule
    Method & 1 & 2 & 3 & 4 & 5 & Avg. L. \\
    \midrule
    MIP ($\pi_{0}$ arch.)$^{*}$ & 85.6 & 70.5 & 57.1 & 48.9 & 40.4 & 3.03 \\
    MIP ($\pi_{0.5}$ arch.)$^{*}$ & 86.7 & 72.1 & 59.6 & 51.2 & 42.4 & 3.12 \\
    $\pi_{0.5}^{*}$ & 90.4 & 78.1 & 67.9 & 61.2 & 54.7 & 3.52 \\
    CF-VLA (full) & \textbf{91.1} & \textbf{80.2} & \textbf{71.8} & \textbf{66.2} & \textbf{57.3} & \textbf{3.67} \\
    \midrule
    CF-VLA w/o Phase II & 88.9 & 77.8 & 67.1 & 59.4 & 50.5 & 3.43 \\
    CF-VLA w/o refine. & 88.4 & 77.5 & 67.7 & 59.3 & 50.6 & 3.44 \\
    CF-VLA w/o var. mod. & 87.7 & 76.8 & 67.3 & 60.2 & 52.5 & 3.45 \\
    CF-VLA w/o Phase I & \textbf{89.2} & \textbf{79.2} & \textbf{69.8} & \textbf{62.5} & \textbf{53.9} & \textbf{3.55} \\
    \bottomrule
  \end{tabular}
\end{table}

\textbf{Results on CALVIN.}
A similar pattern is observed on CALVIN. 
The full model achieves the best long-horizon performance, increasing the average sequence length from 3.12 for MIP ($\pi_{0.5}$ arch.) and 3.52 for $\pi_{0.5}^{*}$ to 3.67. 

Removing Phase I again causes only minor degradation, confirming its role as a training stabilizer. 
In contrast, removing refinement, removing variance modeling, or removing Phase II consistently reduces performance to 3.43--3.45 and leads to larger drops on the more challenging 4- and 5-instruction settings.

This trend supports our core hypothesis: separating global alignment (coarse stage) from local correction (fine stage) is essential for maintaining rollout consistency over long horizons. 
When either stage is removed, the model is forced to compensate for missing structure during inference, leading to increased error accumulation.
\begin{figure}[tb]
  \centering
  \includegraphics[width=\linewidth]{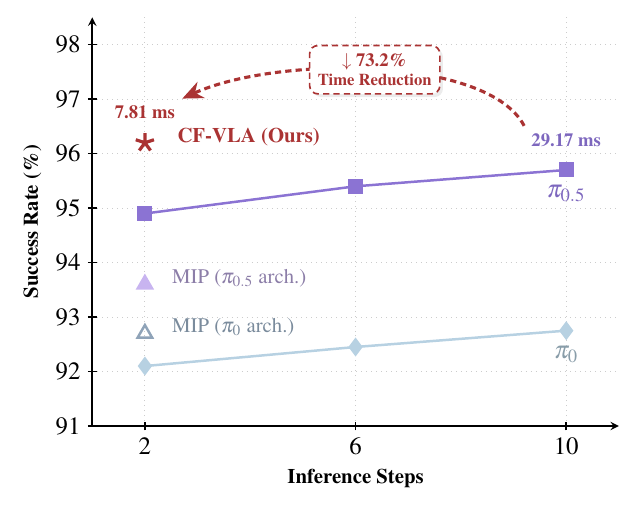}
  \caption{Latency--performance trade-off on LIBERO. We compare average success rate and action sampling latency across methods with different numbers of function evaluations (NFEs). CF-VLA attains a stronger low-NFE operating point, achieving 96.5 average success at 7.81 ms with two function evaluations, compared with 95.7 at 29.17 ms for the reproduced NFE=10 $\pi_{0.5}$ baseline.}
  \Description{A comparison plot on LIBERO showing average success rate versus NFE (inference steps) for representative baselines. CF-VLA reaches 96.5 percent average success at NFE equals 2 with 7.81 ms action sampling latency, while the reproduced NFE equals 10 pi 0.5 baseline reaches 95.7 percent with 29.17 ms. The NFE equals 2 MIP variants and the pi 0 baseline achieve lower average success rates.}
  \label{fig:inference_speed}
\end{figure}
\begin{figure*}[!t]
  \centering
  \includegraphics[width=0.85\linewidth]{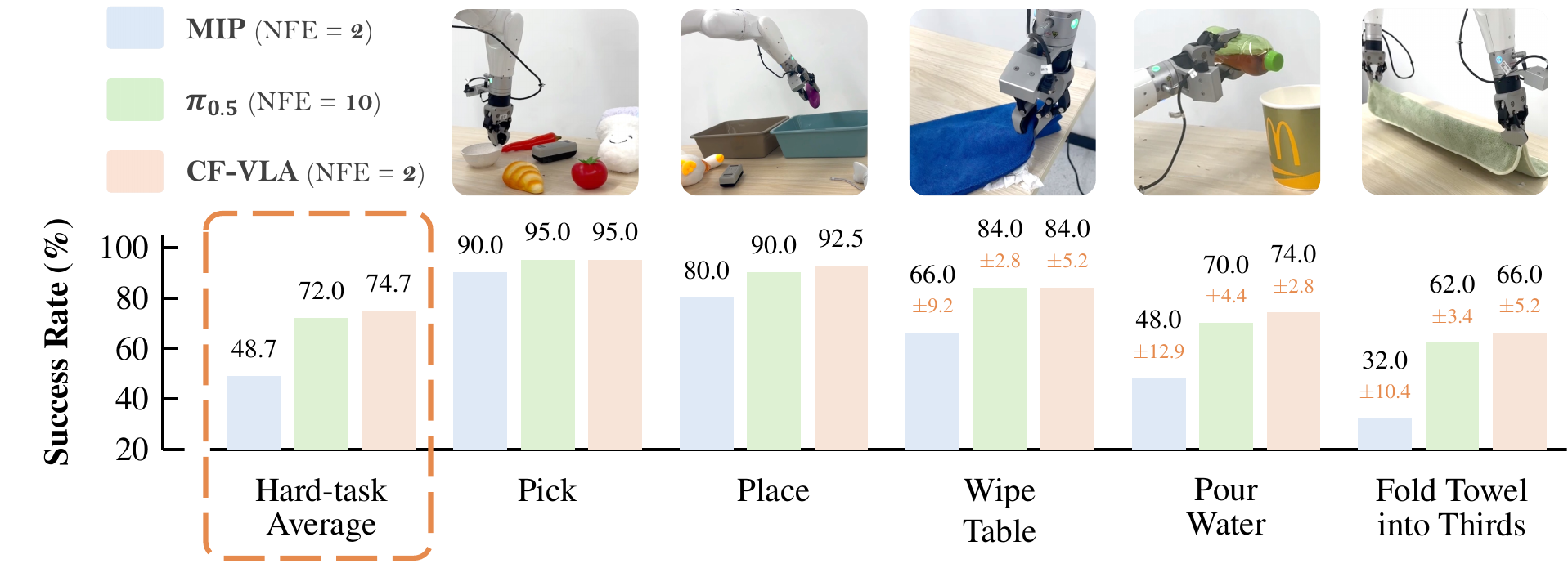}
  \caption{Real-robot results on five representative manipulation tasks. The top panel shows representative task snapshots, and the bottom panel compares success rates of MIP, $\pi_{0.5}$, and CF-VLA, including the hard-task average over Wipe Table, Pour Water, and Fold Towel into Thirds. CF-VLA achieves the highest hard-task average success rate of 74.7\%, compared with 48.7\% for MIP and 72.0\% for $\pi_{0.5}$.}
  \Description{A real-robot results figure with a bar chart and five task photos. The bar chart reports success rates for MIP, pi 0.5, and CF-VLA on Pick X, Put X into Box, Wipe Table, Pour Water, and Fold Towel into Thirds, together with the hard-task average over Wipe Table, Pour Water, and Fold Towel into Thirds. CF-VLA attains the highest hard-task average success rate at 74.7 percent, compared with 72.0 percent for pi 0.5 and 48.7 percent for MIP. The lower panel shows example images for the five tasks.}
  \label{fig:real_robot_results}
\end{figure*}

\textbf{Additional ablations.}
Additional coarse-stage hyperparameter sweeps are provided in the supplementary material.

\subsection{Latency--Performance Frontier}

To complement the NFE-based comparisons, we further analyze the latency--performance trade-off by measuring wall-clock action sampling time, excluding image encoding and KV-cache construction.

As shown in Figure~\ref{fig:inference_speed}, CF-VLA achieves 96.5 average success on LIBERO at NFE=2, surpassing the reproduced NFE=10 $\pi_{0.5}$ baseline (95.7), while reducing action sampling latency from 29.17 ms to 7.81 ms (a 73.2\% reduction). 
At the model level, including input preprocessing, VLM prefix encoding, KV-cache construction, and action sampling, CF-VLA reduces total inference latency from 70.91 ms to 49.55 ms (a 30.1\% reduction), improving the replanning frequency from 14.1 Hz to 20.2 Hz.

Importantly, other NFE=2 baselines in the figure remain significantly below CF-VLA in success rate. 
This indicates that the improvement cannot be attributed to a smaller number of function evaluations alone. 
Instead, it arises from a more effective use of the limited inference budget.

As a result, fewer refinement steps are required, and each step operates in a better-conditioned regime, leading to both higher efficiency and improved action quality.

These results highlight that efficiency in action generation is not solely determined by reducing NFE, but by restructuring how computation is allocated across different stages of the generation process.

\subsection{Real-robot experiments}
\label{sec:real_robot}
\textbf{Real-robot experimental setup.}
We further validate CF-VLA on five representative real-robot manipulation tasks, as shown in Fig.~\ref{fig:real_robot_results}: basic grasping (``Pick X''), pick-and-place (``Put X into Box''), contact-rich wiping (``Wipe Table''), liquid pouring (``Pour Water''), and bimanual deformable-object manipulation (``Fold Towel into Thirds'').

\textbf{Real-robot results.}
As shown in Fig.~\ref{fig:real_robot_results}, CF-VLA achieves the highest hard-task average success rate of 74.7\%, outperforming MIP (48.7\%) by 26.0 points and reaching comparable or slightly better performance than $\pi_{0.5}$ (72.0\%) with a fivefold smaller inference budget. 
On the three harder tasks, an additional five-seed evaluation further confirms this trend, demonstrating that CF-VLA maintains consistent performance across repeated evaluations in these more challenging manipulation settings.
The simpler tasks (``Pick X'' and ``Put X into Box'') mainly serve as sanity checks, while the harder tasks---``Wipe Table,'' ``Pour Water,'' and ``Fold Towel into Thirds''---better test sustained contact, precise trajectory control, and coordinated bimanual execution. On these tasks, CF-VLA matches or improves over the reproduced $\pi_{0.5}$ baseline by 0.0, 4.0, and 4.0 points despite using NFE=2 instead of NFE=10, and improves over MIP under the same NFE=2 budget by 18.0, 26.0, and 34.0 points.

These results are consistent with our core design principle: AP-guided initialization reduces the burden of global alignment during execution, allowing the single refinement step to focus on local correction in contact-rich and long-horizon settings where small deviations can quickly accumulate.
\section{Conclusion}

We presented CF-VLA, a coarse-to-fine framework for efficient flow-based action generation in vision-language-action policies. 
By explicitly separating global alignment from local refinement, CF-VLA restructures the generation process into a two-step procedure consisting of AP-guided initialization and single-step correction.

Across CALVIN, LIBERO, and real-robot benchmarks, CF-VLA achieves a strong balance between performance and efficiency, demonstrating that high-quality action generation does not require extensive iterative refinement. 
Instead, our results show that the structure of the starting point plays a central role in determining both efficiency and robustness.

More broadly, this work suggests a shift in perspective for generative control: rather than allocating computation uniformly along a sampling trajectory, it is more effective to organize computation according to the functional roles of global alignment and local correction. 
We hope this principle can guide the design of future low-latency VLA systems that are deployable in the real world.

\newpage

\bibliographystyle{ACM-Reference-Format}
\bibliography{main}

\end{document}